\documentclass[letterpaper]{article} 
\usepackage{aaai24}  
\usepackage{times}  
\usepackage{helvet}  
\usepackage{courier}  
\usepackage[hyphens]{url}  
\usepackage{graphicx} 
\usepackage{natbib}  
\usepackage{caption} 
\usepackage{algorithm}
\usepackage{algorithmic}
\usepackage{longtable}
\usepackage{booktabs}
\usepackage{newfloat}
\usepackage{listings}
\DeclareCaptionStyle{ruled}{labelfont=normalfont,labelsep=colon,strut=off} 
\floatstyle{ruled}
\newfloat{listing}{tb}{lst}{}
\floatname{listing}{Listing}
\usepackage{xcolor}

\title{Using Adaptive Bandit Experiments to Increase and Investigate Engagement in Mental Health}
\author {
    Harsh Kumar\textsuperscript{\rm 1},
    Tong Li\textsuperscript{\rm 2},
    Jiakai Shi\textsuperscript{\rm 1},
    Ilya Musabirov\textsuperscript{\rm 1},
    Rachel Kornfield\textsuperscript{\rm 3},
    Jonah Meyerhoff\textsuperscript{\rm 3},
    Ananya Bhattacharjee\textsuperscript{\rm 1},
    Chris Karr\textsuperscript{\rm 4},
    Theresa Nguyen\textsuperscript{\rm 5},
    David Mohr\textsuperscript{\rm 3},
    Anna Rafferty\textsuperscript{\rm 6},
    Sofia Villar\textsuperscript{\rm 7},
    Nina Deliu\textsuperscript{\rm 7,8},
    Joseph Jay Williams\textsuperscript{\rm 1}
}
\affiliations {
    \textsuperscript{\rm 1} Department of Computer Science, University of Toronto\\
    \textsuperscript{\rm 2} Department of Statistics, University of Toronto\\
    \textsuperscript{\rm 3} Center for Behavioral Intervention Technologies, Northwestern University\\
    \textsuperscript{\rm 4} Audacious Software\\
    \textsuperscript{\rm 5} Mental Health America\\
    \textsuperscript{\rm 6} Carleton College\\
    \textsuperscript{\rm 7} MRC - Biostatistics Unit, University of Cambridge\\
    \textsuperscript{\rm 8} MEMOTEF Department, Sapienza University of Rome
}

\usepackage{bibentry}

\begin{document}

\maketitle

\begin{abstract}

Digital mental health (DMH) interventions, such as text-message-based lessons and activities, offer immense potential for accessible mental health support. While these interventions can be effective, real-world experimental testing can further enhance their design and impact. Adaptive experimentation, utilizing algorithms like Thompson Sampling for (contextual) multi-armed bandit (MAB) problems, can lead to continuous improvement and personalization. However, it remains unclear when these algorithms can simultaneously increase user experience rewards and facilitate appropriate data collection for social-behavioral scientists to analyze with sufficient statistical confidence. Although a growing body of research addresses the practical and statistical aspects of MAB and other adaptive algorithms, further exploration is needed to assess their impact across diverse real-world contexts. This paper presents a software system developed over two years that allows text-messaging intervention components to be adapted using bandit and other algorithms while collecting data for side-by-side comparison with traditional uniform random non-adaptive experiments. We evaluate the system by deploying a text-message-based DMH intervention to 1100 users, recruited through a large mental health non-profit organization, and share the path forward for deploying this system at scale. This system not only enables applications in mental health but could also serve as a model testbed for adaptive experimentation algorithms in other domains.
\end{abstract}

\section{Introduction}
Enhancing digital mental health interventions can contribute significantly to improved well-being and support in today's society \cite{boardman2011social}. The global prevalence of mental health problems increased by an unprecedented 25\% in 2020, the first year of the COVID-19 pandemic. This increased prevalence and disruptions of traditional mental health care emphasized preexisting gaps in access to support. Before, during and after the pandemic, many cannot access adequate mental health support and are increasingly turning to DMH tools \cite{torous2020digital}. There is an opportunity for AI to help address these challenges and enhance mental health support in various populations \cite{bickman2020improving, kumar2022exploring, kumar2023exploring, kumar2023chatbot}.

Digital Mental Health (DMH) interventions hold promise in extending the availability of support for individuals as they take steps to manage their mental health. Automated messaging is a central component of many DMH interventions as a modality for delivering support and information \cite{bickmore_giorgino_2004}. However, messaging interventions typically deliver the same content across users despite the users' varied needs. This, in turn, reduces the relevance of such messages, as well as compromises engagement and effectiveness \cite{Head2013-uh, bhattacharjee2023investigating}. This is problematic because, while engagement with interventions is not sufficient, it is necessary to benefit users. Moreover, building, optimizing, and evaluating multicomponent interventions is a multi-stage process that is complex and requires data-driven decisions from experts \cite{collins2007multiphase}. 

\begin{figure*}  
  \includegraphics[width=\textwidth]{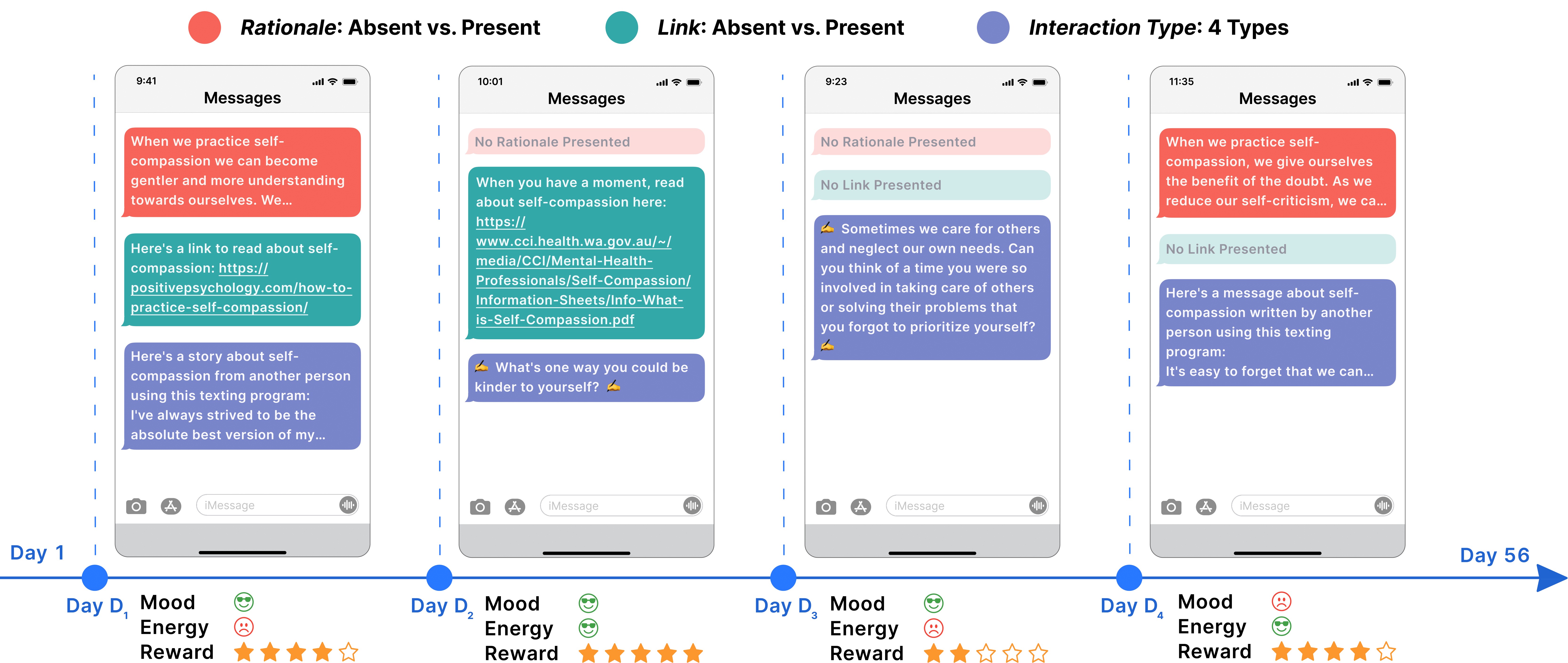}  
  \caption{Schematic representation of an example sequence of messages a user could receive during days D\textsubscript{i}, where D\textsubscript{i} is any random day within the 8-week-long intervention when the user receives a message. \textbf{Mood} and \textbf{Energy} during each day D\textsubscript{i} represent a subset of contexts describing the user during that particular day, and \textbf{Reward} indicates a response from the user to the question \textit{"How helpful were these messages? Reply with a number 1 (not at all helpful) to 5 (very helpful)"} after receiving the messages during the day. The messages are composed of \textit{three} modular components in a 2 (\textit{Rationale}: present vs absent) x 2 (\textit{Link}: present vs absent) x 4 (\textit{Interaction type}: 4 options) factorial design.}  
  \label{fig:teaser}
\end{figure*}

To address these challenges, we developed a system for adaptive experimentation to improve people's engagement with text messages providing a mental health intervention, using Multi-Armed Bandit (MAB) algorithms. The system was built in partnership with Mental Health America (MHA)\footnote{\url{https://mhanational.org/}}, a large non-profit organization dedicated to promoting mental health and preventing mental illness through advocacy, education, research, and services. The system supports an eight-week DMH intervention in which users receive various contents via text messages to help manage their mental health (see Figure \ref{fig:teaser}). To continually test out improvements, our software system instruments components of the text messages to function like intelligent agents that test out different actions (arms), by using MAB algorithms to conduct adaptive experiments.

The purpose of using this system is twofold. 
\begin{enumerate}
    \item First, we are able to run algorithms in a real-world mental health setting to optimize efficacy, as has been done in product development.
    \item Second, this system allows us to understand how well the results of such algorithms can be analyzed by clinical scientists with statistical rigor. 
\end{enumerate}  
For example, consider an Action (experimental) variable that is testing whether providing a rationale for using a particular psychological strategy impacts the Reward (outcome) -- a measure of message helpfulness. The impact of the Action variable on the Reward may depend on contextual variables, such as self-reported Mood. Domain scientists, like clinical psychology researchers, may want to analyze such data to understand how the Action variable impacts the Reward, and how that impact is mediated by covariates or contextual variables like Mood.

Traditional experiments with uniform random assignment generate data that can be used to provide answers to these questions. However, traditional experiments do not dynamically provide a better user experience by rapidly using existing data to inform what a user will receive in the future \cite{bickman2020improving}.  
Using the most fundamental formulation of exploration-exploitation tradeoffs, (contextual) MAB algorithms allow for such a dynamic, adaptive experimental design. This can have the benefit of increasing the probability that users are assigned to actions that work best for them, as well as the benefit of decreasing the probability that users receive burdensome or ineffective actions. 

However, there might be a trade-off between collecting data to answer domain scientist questions and using data to try to optimize for users \cite{yao2021power}. For example, research has suggested that there may be biases in estimates of arm means, increased false positive rates, and reduced statistical power \cite{Williams2021-ll, postdiff}. Our system not only enables the use of contextual MAB algorithms to optimize for users, but also provides a testbed for evaluating how different scenarios can impact false positive rates, statistical power, and reward maximization. 


\section{Related Work}
\subsection{Digital Mental Health and Text-message based interventions}
Digital mental health (DMH) interventions provide a solution for narrowing the service provision gap, wherein the need for mental health support far outweighs the availability of traditional care \cite{Gan2021-yd}. By utilizing tools like computers and smartphones, psychological treatment can become widely accessible. The majority of Americans own smartphones \cite{rainie_chapter_2015}, and DMH interventions using this modality allow users to customize how they wish to receive and engage with mental health support. Previous work has found DMH interventions for depression to be effective across randomized control trials \cite{firth_efficacy_2017}. However, user engagement levels with DMH interventions are low, barring adoption into the real world \cite{Gan2021-yd}. 

Text messaging as a modality has proven to be effective in promoting behavior change. Using automated messaging, short content that is repeatedly delivered throughout the day can lay the foundation for a larger behavior change process \cite{Abroms2015-ka}. Further, automated messaging has been successful in promoting and maintaining positive mental health \cite{armanasco2017preventive}.

\subsection{Adaptive Experiments and Multi-Armed Bandit Algorithms}
Contextual MAB algorithms are one solution for adaptive experiments. These algorithms base the probability of assignment on previous data, ensuring that more users are exposed to more effective conditions, and fewer users are exposed to less effective conditions \cite{li2010contextual}. Bandit algorithms have been extensively applied in real-world settings to tackle the exploration-exploitation dilemma in sequential decision-making \cite{fouche2019scaling,zhan2021off}. However, there are issues that arise from MAB algorithms that complicate data analysis. Primarily, these algorithms have been shown to produce measurement errors, increases in false positive rates, and decreases in statistical power \cite{russo_2020, xiang2022multi}. These hindrances work directly against the objective of the algorithm by decreasing the chance that effective conditions will be identified and increasing the probability that unhelpful conditions will be deployed \cite{rafferty2019statistical}. Therefore, it may be helpful to use systems that enable comparisons of different algorithms, such that researchers can make informed decisions about these statistical trade-offs. 

\section{Design of Intervention}
\label{section:intervention}
A series of design workshops were conducted with non-treatment seeking participants \cite{Kornfield2022-ul, meyerhoff2022meeting, kornfield2023text} to generate content which could be tailored according to users' preferences, backgrounds, and contexts. Detailed design of the overall study is reported in a separate paper \cite{meyerhoff2022system}.


In the next section, we deep dive into a particular component of the intervention to show as an example how contextual multi-armed bandit algorithms are used throughout the numerous experimentation points.

\subsection{Adaptive Components in DMH Intervention}
\label{section:modular}

Modular dialogues are brief, self-contained interactions that support a single psychological strategy, and can be delivered at one contact point without the need to reference messages earlier or later in the day. Our software system enables embedded adaptive experiments at a number of \textit{intelligent} decision points in these modular dialogues. 

Figure \ref{fig:teaser} illustrates an example of message sequences a participant may receive during an 8-week intervention to practice self-compassion. The messages contain three modular components (decision points) in a factorial design: \textit{Rationale} (present vs. absent), \textit{Link} (present vs. absent), and \textit{Interaction type} (four options). Our software system enables adaptive experiments at various intelligent decision points in these modular dialogues.

We use contextual multi-armed bandit algorithms throughout numerous experimentation points to adapt the assignment of arms based on the collected data. The system captures various contextual variables to tailor messages according to users, such as user profiles, preferences, and interactions with the system during the intervention (Table \ref{table:context}). The coding of these variables can be adjusted as needed.

Rewards in the form of message ratings and link clicks are used to optimize content quickly. For instance, at the end of each message sequence, users rate the helpfulness of the messages on a scale of 1 to 5. The bandit algorithm also considers whether users click the embedded hyperlinks in the messages.

The bandit algorithm adapts the assignment of action variables based on the collected data. For example, it decides whether to provide a Rationale for introducing a psychological strategy, include a Link to web content, or select one of the four Interaction types. The system is designed to capture rewards of varying natures to enable rapid prototyping and testing of different rewards for different contexts.

\begin{table}[]
\setlength{\tabcolsep}{5pt}
\renewcommand{\arraystretch}{1}
\resizebox{1\columnwidth}{!}{\begin{tabular}{p{0.08\textwidth}p{0.085\textwidth}p{0.28\textwidth}}
\toprule \textbf{Variable} & \textbf{Type} & \textbf{Description}\\ 
\midrule
Mood \& Energy & Binary  & Takes 0 when users report Low or Medium levels, and 1 for High levels\\ 
K10 & Ordinal & Based off of the Kessler Psychological Distress Scale. Bounded between 1 and 4. \cite{ACI} \\ 
Recent Activity Last 48 hours & Binary & Takes 0 when user is inactive in the last 48 hours of receiving the message, and 1 if user interacts in any way with the system (e.g. types something, clicks on any link, etc). \\ \hline
\end{tabular}}
\caption{Some user contexts captured by the system (out of 12 total contextual variables).}
\label{table:context}
\end{table}

\subsection{Algorithms for Adaptive Experimentation}
\label{section:algorithm}

The system supports a range of algorithms for adaptive experimentation, which dynamically update the assignment probability of each intervention based on the observed data. Below we describe the application of one of the algorithms.
\subsubsection{Problem Formulation}
Assuming an experiment over a horizon of size $T$, the problem is to choose a sequence of $T$ actions $\{a_t\}_{t=1,\dots, T}$ that maximize the expected cumulative reward over time $E[\sum_{t=1}^{T}r_{a_t}(t)]$, with $r_{a_t}(t)$ being the reward associated to action $a_t$ at time $t$.
One way to formalize the problem of dynamic decision-making about what type of message to deliver \textit{to whom} is by using a \textit{contextual} MAB problem~\cite{li2010contextual, agrawal2013}. In contextual MABs, the reward is conceived as a function of users' contexts, in addition to allocated arms, so it allows \textit{personalization} as well. 
We formulate each dimension of the messaging protocol (framing, hyperlink, interaction type) as a separate intelligent decision point, so we solve three contextual bandit problems. Here, we illustrate the setup of a popular MAB strategy based on contextual linear TS~\cite{agrawal2013}. 

\subsubsection{Algorithm}
Suppose that the likelihood of the observed reward $r_i (t)$ of arm $i$ at step $t$, given a context vector $b_i(t) \in \mathrm{R}^d$ and the unknown parameter vector $\mu \in \mathrm{R}^d$, is given by a Gaussian model with mean $b_i(t)\mu$ and standard deviation $v=R\sqrt{\frac{24}{\epsilon}d\ln{(\frac{1}{\delta})}}$ with $\epsilon \in (0,1)$. Denoting with $B(t)=I_d + \sum_{\tau}^{t-1} b_{a_\tau}(\tau)b_{a_\tau}(\tau)^T$ and with $\hat{\mu}(t) = B(t)^{-1}\left(\sum_{\tau}^{t-1} b_{a_\tau}(\tau)r_{a_\tau}(\tau)\right)$, if the prior at step $t$ is $\mathcal{N}(\hat{\mu}(t), v^2 B(t)^{-1})$, then the posterior at $t+1$ is given by $\mathcal{N}(\hat{\mu}(t+1), v^2 B(t+1)^{-1})$. In practice, at each step $t$, a sample $\tilde{\mu}(t)$ is be generated from $\mathcal{N}(\hat{\mu}(t), v^2B(t)^{-1})$, and the $i$-th arm which maximizes $b_i(t)^T\tilde{\mu}(t)$ is played. 



\section{Evaluation of System}
\label{section:simulation}


After extensive internal testing, we deployed this system with 50 participants (recruited from the MHA website) to undergo the eight-week intervention. This helped us design simulation scenarios and make adjustments to ensure data is being appropriately collected, and that the algorithms are adequately adapting the experiments. 
Based on data from pilot deployments, we illustrate the following simulated scenarios. The code to reproduce the following analysis is made publicly available\footnote{\url{https://github.com/harsh-kumar9/bandit_simulation}}.  

\subsection{Simulation settings}
We generate each user's context randomly (with all possible values being equally likely). To resemble our real-world scenario, where the users' feedback rating has 5 levels (see Figure \ref{fig:teaser}), we simulate reward values from $ \mathcal{R} = \{0, 0.25, 0.5, 0.75, 1\}$. The generating process involves first, generating a raw reward from a Normal distribution, and then rounding it to the closest value in $\mathcal{R}$. We consider 'Rationale' framing (arm $i=1$) versus 'No Rationale' framing (arm $i=0$) arms, and context.

\textbf{Scenario 1: No arm difference} In this case, the raw-reward $r_i(t)$ follows the same distribution for both arms and does not depend on the context:
$$
r_i(t) \sim N\left(0.5,(1/6)^2\right), \quad i=0,1.
$$

\textbf{Scenario 2: Substantial arm difference} Here, we assume that arm $i=1$ works better than the arm  $i=0$: 
$$r_i(t) \sim N\left(0.5+1/8\times i,(1/6)^2\right), \quad i=0,1.$$

\textbf{Scenario 3: Optimal arm changes based on the context} We assume that the reward for each arm $i=0,1$ changes based on an interaction between arm and context. Denoting the interaction with $*$, this is given by: 
$$
r_i(t) \sim N\left(0.5+\frac{3}{8} \times i-\frac{1}{4} \times m_t-\frac{5}{8} \times i * m_t, (\frac{1}{6})^2\right)
$$
where $m_t$ is the context variable assuming value 1 if the $t$-th participant has good mood, and 0 otherwise. According to our setting, when a participant has a high mood, the arm 'Rationale' has a higher expected raw reward (0.875 with 'Rationale' versus 0.5 with 'No Rationale'), while the opposite is true for participants with a low mood (0 with 'Rationale' versus 0.25 with 'No Rationale'). 

\subsection{Simulation analysis}
For statistical analysis, the data collected can be used to run simulations of the algorithm's behavior if an experiment was run thousands of times. We present the Reward, False Positive Rate (FPR; the probability of a statistical test to incorrectly report an arm difference, when one does not exist), and Statistical Power (the probability of a test to correctly report an arm difference, when one exists). The hypothesis tests related to FPR and Power are conducted by sampling from the joint posterior distribution of the parameters of interest and estimating the 95$\%$ Confidence Intervals. Compared with traditional uniform random experimentation, these reveal the extent to which an algorithm's adaptive data collection increases, reduces, or has no effect on reliably detecting different kinds of effect - like whether or how much better an arm is or which arm is best based on a contextual variable.

\begin{figure}[ht]
    \includegraphics[width=0.485\textwidth]{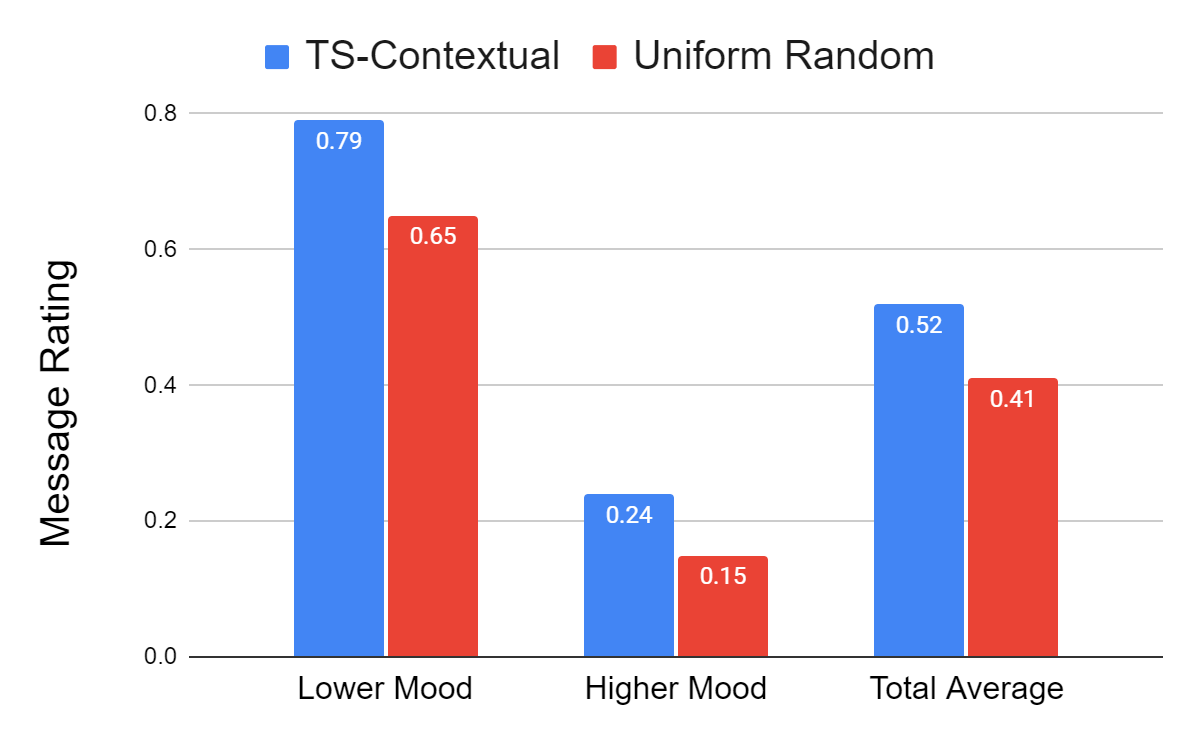}
    \caption{Average rewards using Thompson Sampling for Contextual Bandits (Contextual TS) versus Uniform Random in different cases. The first pair of bars compares the reward in the group of participants having \textit{low mood}, the second pair compares in \textit{high mood} group, and the last pair takes the average among all participants.}
    \label{fig:mood}
\end{figure}

From Figure \ref{fig:mood}, we see that Contextual TS can adapt to the data and lead to a higher reward than Uniform Random when there is a difference in the expected arm rewards. For people in a good mood, Contextual TS assigns most of them to the 'Rationale' arm, and gets 0.14 more average reward than Uniform Random; for people in a bad mood, Contextual TS assigns the arm with 'No Rationale' more frequently.

\label{system}

Table \ref{table:fpr} shows the FPR results in our first simulation setting (Scenario 1). The FPR for Contextual TS is almost as good as the Uniform Random algorithm when the sample size is N = 100, and it increased to 0.07 when the sample size increases to N = 1000. This is because we chose a relatively uninformative prior, which forces Contextual TS to explore more and reduces FPR. Increasing the sample size, the effect of the prior is reduced, and an exploitative algorithm like TS may allocate the two arms unevenly, impacting statistical inference ability. However, a value of 0.07 may be regarded as a good trade-off given the higher average reward. This highlights how one could further reduce the FPR by using a wider confidence interval, or choosing a more uninformative prior – making the choice to give up some reward.

Table \ref{table:power} summarizes the results for Power in simulation Scenarios 2 and 3. In Scenario 2, when the sample size is small (N = 100), the Contextual TS algorithm has 15$\%$ less Power than Uniform Random. However, when the sample size is fairly large, Power is not much of a concern, as both approaches produce Power close to 1. In Scenario 3, where the reward is a more complicated function of 'Rationale' and 'Mood', the Contextual TS algorithm has good discovery rates on detecting effects in terms of 'Rationale' and the interacting effect 'Rationale $*$ Mood'. The only drawback for Contextual TS, in this case, is its low Power on the single effect on 'Mood'. This occurs because Contextual TS assigns most people with a good mood to the arm 'Rationale' and those with a bad mood  to 'No Rationale' arm, increasing Reward for both groups, whereas UR more accurately evaluates the effect of Mood across both arms. 
\begin{table}[]
\setlength{\tabcolsep}{5pt}
\renewcommand{\arraystretch}{1}
\begin{center}
\resizebox{0.8\columnwidth}{!}{\begin{tabular}{lcccc}
\toprule
               & &N=100            & &N=1000           \\ \hline
Contextual TS  & &0.03           & &0.07           \\
Uniform Random & &0.04           & &0.04           \\ \hline
\end{tabular}}
\caption{False Positive Rate (FPR) in simulation Scenario 1 where there are no differences between the arms.}
\label{table:fpr}
\end{center}
\end{table}
The data can also be used to understand the detailed behavior of an algorithm over time, and reveal violations of assumptions and interesting extensions based on the structure of the data, like unforeseen subgroups, non-stationarity, or correlations between observations. 
\begin{table}[]
\setlength{\tabcolsep}{5pt}
\renewcommand{\arraystretch}{1}
\begin{center}
\resizebox{0.8\columnwidth}{!}{\begin{tabular}{llcccc}
\toprule
                   &                 & & N=100        &    & N=1000           \\ 
\multicolumn{6}{c}{\textbf{Scenario 2}}                                         \\ \midrule
{\bfseries Contextual TS}&&&&&\tabularnewline
~~Rationale&&&0.72&&0.97\tabularnewline
{\bfseries Uniform Random}&&&&&\tabularnewline
~~Rationale&&&0.87&&1.00\tabularnewline
\multicolumn{6}{c}{\textbf{Scenario 3}}                                         \\ \midrule
{\bfseries Contextual TS}&&&&&\tabularnewline
~~Rationale & && 0.92& & 1.00\tabularnewline
~~Mood & && 0.41 && 0.47\tabularnewline
~~Rationale * Mood & && 1.00 && 1.00\tabularnewline
{\bfseries Uniform Random}&&&&&\tabularnewline
~~Rationale & && 1.00 && 1.00\tabularnewline
~~Mood & && 0.95 && 1.00\tabularnewline
~~Rationale * Mood & && 1.00 && 1.00\tabularnewline
\bottomrule
\end{tabular}}
\caption{Calculation of Power for Scenario 2 (substantial difference between the arms) and 3 (optimal arm changes based on context).}
\label{table:power}
\end{center}
\end{table}

These simulations are an important part of the intervention design process. First, they help evaluate the effectiveness of adaptive algorithms for exploration-exploitation, such as MAB-based strategies, in terms of how they balance reward and rigor of statistical analysis. Second, when tuned for the specific intervention design problem, they serve as a foundation for fine-grained intervention design discussion between domain scientists, developers, and machine learning specialists, allowing exploration of multiple alternative adaptive experimentation scenarios in mental health even before data collection.


\subsection{Real-world Deployment}
To evaluate the system and the intervention experimentally, we deployed the system with 1100 users recruited through an online ad on the MHA website in multiple batches. Each user consented to enroll in the 8-week program, where they received \textit{modular} messages (visually represented in Figure \ref{fig:teaser}) 2-3 times a week, as part of a larger text-message-based intervention. 
We structure the analysis around two factors of the intervention, \textit{Rationale} (present vs absent) and \textit{Link} (present vs absent).

\begin{table}[htbp]
  \centering

\begin{tabular}{lccc}
\toprule
& \multicolumn{2}{c}{Provided Response} &  \\ 
\cmidrule(lr){2-3}
& No & Yes & Total \\ 
\midrule &  &  &  \\ 
Link Present & 610 (78\%) & 171 (22\%) & 781 (100\%) \\ 
Link Absent & 628 (80\%) & 160 (20\%) & 788 (100\%) \\ 
Total & 1,238 (79\%) & 331 (21\%) & 1,569 (100\%) \\ 
\bottomrule
\end{tabular}
\caption{Response rate by arm for the factor "Link". We do not observe large differences between arms in terms of response rate (whether the user responds to the message or not).}
\label{table:expt_responce}
\end{table}

\subsubsection{Engagement}
For 8,521 total arm assignments, we received 813 ratings (9.54\% engagement rate), contributed by 230 unique users (20.9\% of total users). Overall engagement is low, as is characteristic of DMH interventions \cite{Gan2021-yd}. To check if the engagement depends on the experimental arms, we analyzed the rate by arms (see Table \ref{table:expt_responce} for the Link arm), observing no large differences. 
\begin{table}
\resizebox{\columnwidth}{!}{\begin{tabular}{lcccc}
\toprule
 & \multicolumn{2}{c}{\textbf{Contextual TS}} & \multicolumn{2}{c}{\textbf{Uniform Random}} \\ 
\cmidrule(lr){2-3} \cmidrule(lr){4-5}
\textbf{Factor} & \textbf{Present} & \textbf{Absent} & \textbf{Present} & \textbf{Absent} \\ 
\midrule
& N = 232 & N = 175 & N = 202 &  N = 204 \\ 

Link & 0.790 (0.018)\textsuperscript{1} & 0.716 (0.024)\textsuperscript{1} & 0.719 (0.021)\textsuperscript{1} & 0.640 (0.024)\textsuperscript{1} \\ 
\midrule
 & N = 282 & N = 167 & N = 192 &  N = 171 \\ 

Rationale & 0.736 (0.018)\textsuperscript{1} & 0.728 (0.025)\textsuperscript{1} & 0.710 (0.022)\textsuperscript{1} & 0.703 (0.024)\textsuperscript{1} \\ 
\bottomrule
\textsuperscript{1}Mean (SEM)\\
\end{tabular}}
\caption{Overall summary including Mean and Standard Error of Mean (SEM) of the rewards collected for the two factors (Link and Rationale).}
\label{table:expt_summary}
\end{table}




\subsubsection{Efficiency of Policies}

Table \ref{table:expt_summary} shows the summary of data for the different factors and arms of the experiment. We observe an increase in Mean rewards for both levels of each factor for Contextual TS, compared to Uniform Random assignment.

\begin{figure}[htp]
\centering
\begin{minipage}{\linewidth}
\includegraphics[width=1\textwidth]{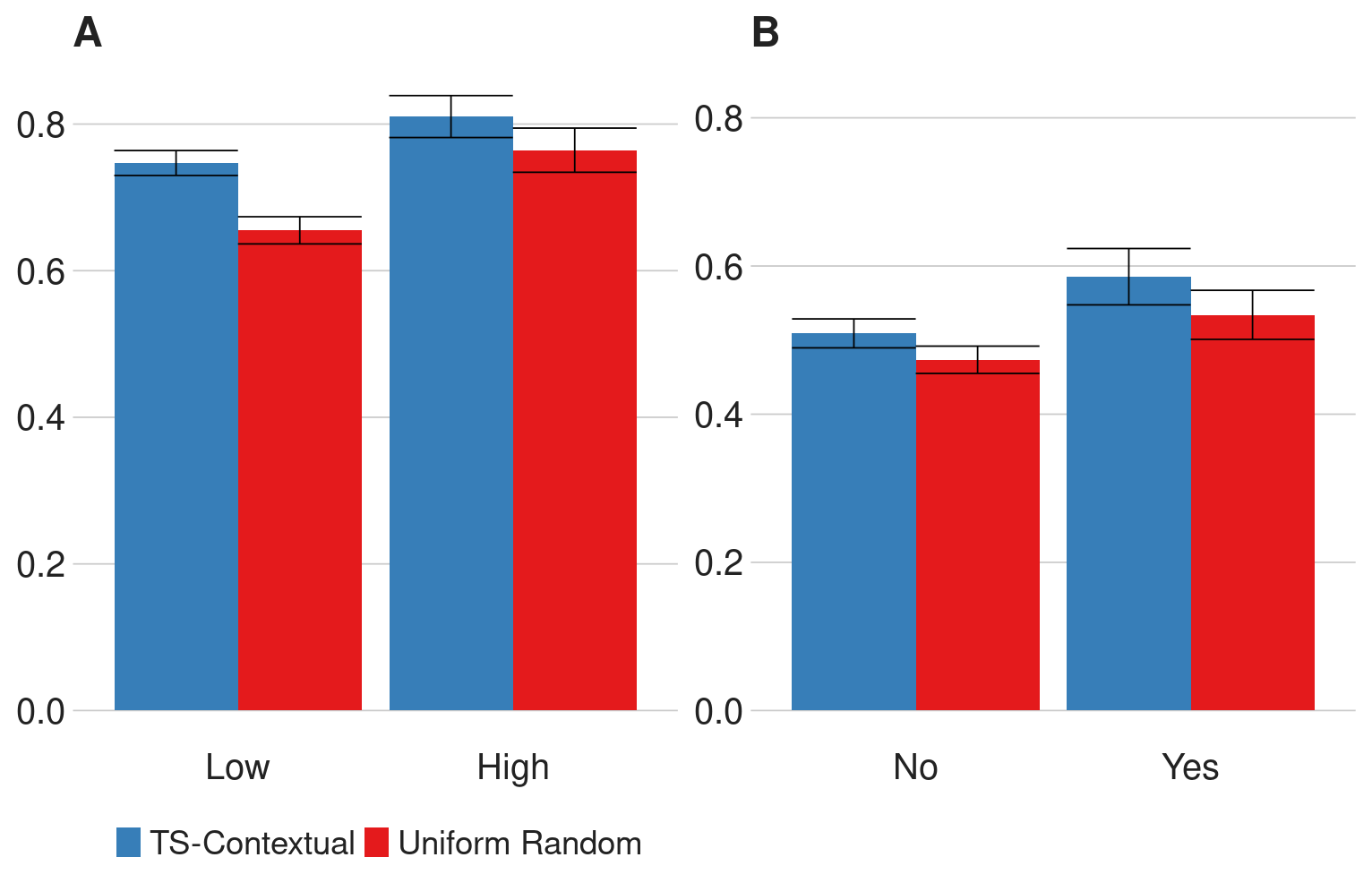}
\end{minipage}
\caption{Average reward (rating of 1 to 5 scaled) using Contextual TS versus Uniform Random for "Link" rating for different levels of contextual variables. Figure A shows the distribution for contextual variable \textit{Mood} (Low vs High). Number of participants (N) from left to right is [322, 316, 83, 87]. Figure B shows the distribution for \textit{Activity in last 48 hours} (Yes vs No). N from left to right are [67, 75, 338, 329].}
\label{fig:contextual_expt}
\end{figure}


\subsubsection{Contextual Effects} 
\begin{figure*}[h]
    \includegraphics[width=\linewidth]{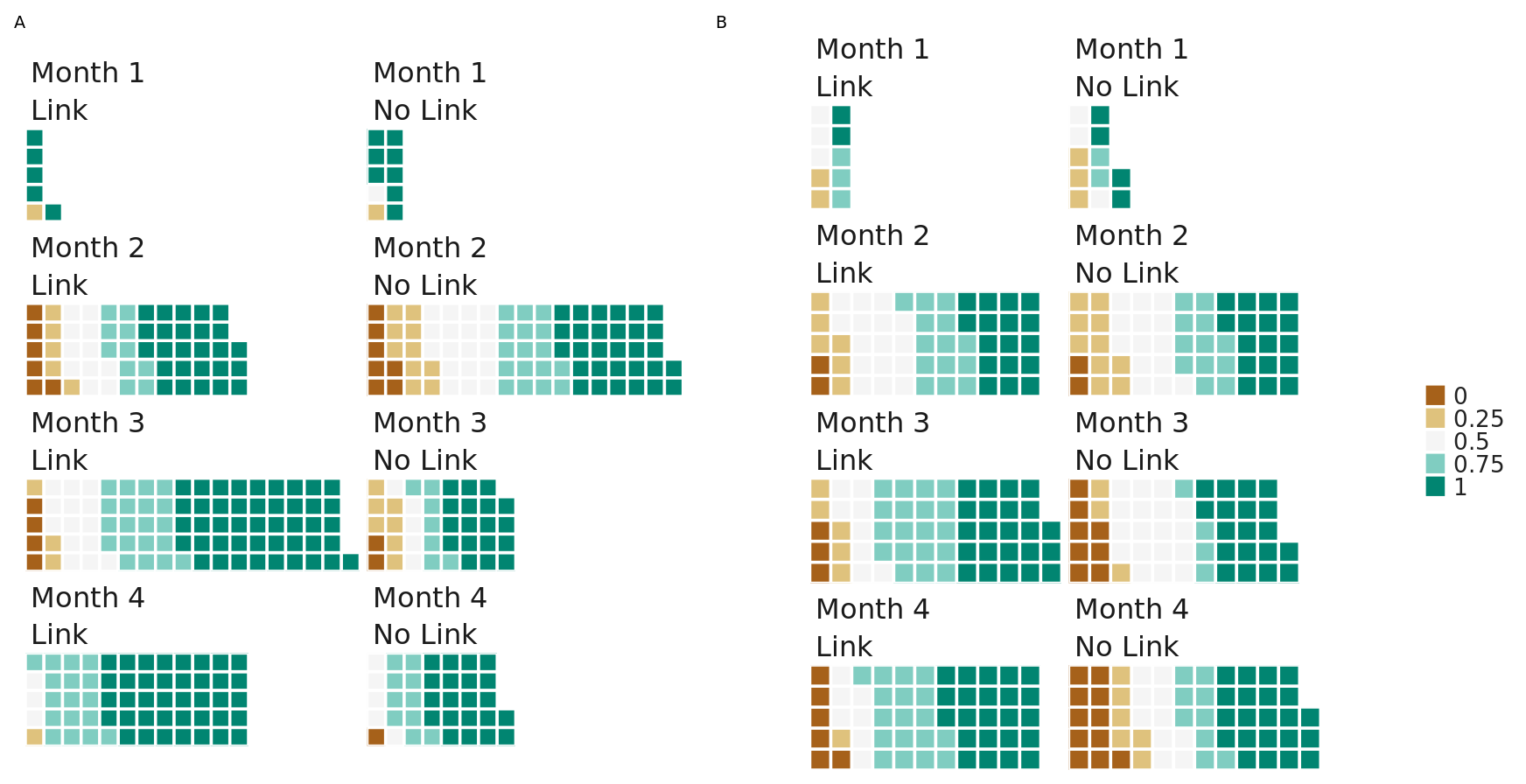}
    \caption{Arm allocation dynamics for Contextual Thompson Sampling (left) vs Uniform Random (right). On both parts of the graph, columns represent arms, and grid rows artificially split experiments by approximately one-month periods, allowing to compare arm allocation in different stages of the experiment. Each small square is one reward we receive with a fill color representing how helpful the participant found the message. On the right, participants are assigned to a traditional experiment aiming for a constant 50-50 split. The "No Link" condition tends to yield lower rewards (represented by brownish squares). On the left, adaptivity is displayed. In the first period, there was a marginally better response for "No Link," leading to some allocations to this arm in the second period. However, the algorithm was able to adjust based on responses, consistently allocating more interactions to the "Link" arm in months 3 and 4.}
    \label{fig:ctsvsur}
\end{figure*}

Checks on potential contextual effects are difficult in the early stages due to higher power requirements for detecting interactions, but they are still important. Here we present two summary analyses for promising contextual variables. 
In Figure \ref{fig:contextual_expt}, we can see that Contextual TS with Mood as a contextual variable was able to achieve a slightly higher average reward compared to Uniform Random. This corresponds to the results of one of the simulation scenarios we developed in the earlier stage of the analysis (Figure \ref{fig:mood}), contributing to the reinforcement of our hypothesis about the role of mood in moderating reactions. Figure \ref{fig:ctsvsur} shows how Contextual TS adapts the assignments for Link vs No Link decisions based on users' interactions.


For the Recent Activity variable, the effect is less pronounced for not recently active participants but observed for those recently active. One design decision to consider based on these results might be to revise the 48h cut-off or make it adaptive.


\section{Discussion and Conclusion}

A key insight in the design of our system was to build flexibility in how algorithms are applied. First, in open source testbed where algorithms can be directly uploaded and used in production within days\footnote{\url{https://github.com/Intelligent-Adaptive-Interventions-Lab/mooclet-engine}}. Second, in allowing for algorithms to flexibly adjust which reward and contextual variables are being used and how, with rapid redeployment. For example, the current reward for the \textit{Rationale} bandit problem is the message rating provided by the user. However, other variables could be used as reward to optimize the messages. The system also allows for new arms to be added by the social-behavioural science design team, as the adaptive experiments reveal that some arms are less effective, as we scale to 5000 users.

Our system provides a testbed for optimizing user experience in mental health while also facilitating critical analysis of algorithms, particularly understanding how algorithms balance reward maximization with data collection for statistical analysis. This places more emphasis on supporting domain scientists to answer socio-behavioral research questions and to draw conclusions that can be generalized to many future users, which is increasingly being emphasized in applications of bandit algorithms \cite{yao2021power}. There is a growing understanding in the behavioral science and policy communities that behavioral interventions require taking into account the heterogeneity of treatment effects \cite{bryan2021behavioural}, including adapting to the contexts on the participant or situational level. This is also relevant for vulnerable and underprivileged populations, providing intervention designers with the tools to break the vicious circle of optimizing for the average participant. Future work can look into incorporating fairness metrics to evaluate the fairness of these algorithms when applied in a mental health setting \cite{wang2022survey, joseph2016fairness, huang2022achieving}.

We showed how the algorithms trade-off optimization of reward (by giving the better arm on average or providing a personalized user experience) while collecting data that enables statistical inference. These just begin to scratch the surface of the complex real-world scenarios our system opens doors for future investigation of - rewards that are dependent on multiple interactions between different action variables, non-stationary, between contextual variables, when data are missing in different ways, and with repeated observations from the same person.


\bibliography{aaai24}

\begin{thebibliography}{36}
\providecommand{\natexlab}[1]{#1}

\bibitem[{ACI(2001)}]{ACI}
 2001.
\newblock https://aci.health.nsw.gov.au.

\bibitem[{Abroms et~al.(2015)Abroms, Whittaker, Free, Mendel Van~Alstyne, and Schindler-Ruwisch}]{Abroms2015-ka}
Abroms, L.~C.; Whittaker, R.; Free, C.; Mendel Van~Alstyne, J.; and Schindler-Ruwisch, J.~M. 2015.
\newblock Developing and pretesting a text messaging program for health behavior change: Recommended steps.
\newblock \emph{JMIR MHealth UHealth}, 3(4): e107.

\bibitem[{Agrawal and Goyal(2013)}]{agrawal2013}
Agrawal, S.; and Goyal, N. 2013.
\newblock Thompson Sampling for Contextual Bandits with Linear Payoffs.
\newblock In \emph{International Conference on Machine Learning}. JMLR:W\&CP.

\bibitem[{Armanasco et~al.(2017)Armanasco, Miller, Fjeldsoe, and Marshall}]{armanasco2017preventive}
Armanasco, A.~A.; Miller, Y.~D.; Fjeldsoe, B.~S.; and Marshall, A.~L. 2017.
\newblock Preventive health behavior change text message interventions: a meta-analysis.
\newblock \emph{American journal of preventive medicine}, 52(3): 391--402.

\bibitem[{Bhattacharjee et~al.(2023)Bhattacharjee, Williams, Meyerhoff, Kumar, Mariakakis, and Kornfield}]{bhattacharjee2023investigating}
Bhattacharjee, A.; Williams, J.~J.; Meyerhoff, J.; Kumar, H.; Mariakakis, A.; and Kornfield, R. 2023.
\newblock Investigating the Role of Context in the Delivery of Text Messages for Supporting Psychological Wellbeing.
\newblock In \emph{Proceedings of the 2023 CHI Conference on Human Factors in Computing Systems}, 1--19.

\bibitem[{Bickman(2020)}]{bickman2020improving}
Bickman, L. 2020.
\newblock Improving mental health services: A 50-year journey from randomized experiments to artificial intelligence and precision mental health.
\newblock \emph{Administration and Policy in Mental Health and Mental Health Services Research}, 47(5): 795--843.

\bibitem[{Bickmore and Giorgino(2004)}]{bickmore_giorgino_2004}
Bickmore, T.; and Giorgino, T. 2004.
\newblock \emph{Some Novel Aspects of Health Communication from a Dialogue Systems Perspective}.

\bibitem[{Boardman(2011)}]{boardman2011social}
Boardman, J. 2011.
\newblock Social exclusion and mental health--how people with mental health problems are disadvantaged: an overview.
\newblock \emph{Mental Health and Social Inclusion}, 15(3): 112--121.

\bibitem[{Bryan, Tipton, and Yeager(2021)}]{bryan2021behavioural}
Bryan, C.~J.; Tipton, E.; and Yeager, D.~S. 2021.
\newblock Behavioural science is unlikely to change the world without a heterogeneity revolution.
\newblock \emph{Nature human behaviour}, 5(8): 980--989.

\bibitem[{Collins, Murphy, and Strecher(2007)}]{collins2007multiphase}
Collins, L.~M.; Murphy, S.~A.; and Strecher, V. 2007.
\newblock The multiphase optimization strategy (MOST) and the sequential multiple assignment randomized trial (SMART): new methods for more potent eHealth interventions.
\newblock \emph{American journal of preventive medicine}, 32(5): S112--S118.

\bibitem[{Firth et~al.(2017)Firth, Torous, Nicholas, Carney, Pratap, Rosenbaum, and Sarris}]{firth_efficacy_2017}
Firth, J.; Torous, J.; Nicholas, J.; Carney, R.; Pratap, A.; Rosenbaum, S.; and Sarris, J. 2017.
\newblock The efficacy of smartphone-based mental health interventions for depressive symptoms: a meta-analysis of randomized controlled trials.
\newblock \emph{World Psychiatry}, 16(3): 287--298.

\bibitem[{Fouch{\'e}, Komiyama, and B{\"o}hm(2019)}]{fouche2019scaling}
Fouch{\'e}, E.; Komiyama, J.; and B{\"o}hm, K. 2019.
\newblock Scaling multi-armed bandit algorithms.
\newblock In \emph{Proceedings of the 25th ACM SIGKDD International Conference on Knowledge Discovery \& Data Mining}, 1449--1459.

\bibitem[{Gan et~al.(2021)Gan, McGillivray, Han, Christensen, and Torok}]{Gan2021-yd}
Gan, D. Z.~Q.; McGillivray, L.; Han, J.; Christensen, H.; and Torok, M. 2021.
\newblock Effect of engagement with digital interventions on mental health outcomes: A systematic review and meta-analysis.
\newblock \emph{Front Digit Health}, 3: 764079.

\bibitem[{Head et~al.(2013)Head, Noar, Iannarino, and Grant~Harrington}]{Head2013-uh}
Head, K.~J.; Noar, S.~M.; Iannarino, N.~T.; and Grant~Harrington, N. 2013.
\newblock Efficacy of text messaging-based interventions for health promotion: a meta-analysis.
\newblock \emph{Soc. Sci. Med.}, 97: 41--48.

\bibitem[{Huang et~al.(2022)Huang, Labille, Wu, Lee, and Heffernan}]{huang2022achieving}
Huang, W.; Labille, K.; Wu, X.; Lee, D.; and Heffernan, N. 2022.
\newblock Achieving user-side fairness in contextual bandits.
\newblock \emph{Human-Centric Intelligent Systems}, 1--14.

\bibitem[{Joseph et~al.(2016)Joseph, Kearns, Morgenstern, and Roth}]{joseph2016fairness}
Joseph, M.; Kearns, M.; Morgenstern, J.~H.; and Roth, A. 2016.
\newblock Fairness in learning: Classic and contextual bandits.
\newblock \emph{Advances in neural information processing systems}, 29.

\bibitem[{Kornfield et~al.(2022)Kornfield, Meyerhoff, Studd, Bhattacharjee, Williams, Reddy, and Mohr}]{Kornfield2022-ul}
Kornfield, R.; Meyerhoff, J.; Studd, H.; Bhattacharjee, A.; Williams, J.~J.; Reddy, M.; and Mohr, D.~C. 2022.
\newblock Meeting users where they are: User-centered design of an automated text messaging tool to support the mental health of young adults.
\newblock In \emph{{CHI} Conference on Human Factors in Computing Systems}. New York, NY, USA: ACM.

\bibitem[{Kornfield et~al.(2023)Kornfield, Stamatis, Bhattacharjee, Pang, Nguyen, Williams, Kumar, Popowski, Beltzer, Karr et~al.}]{kornfield2023text}
Kornfield, R.; Stamatis, C.~A.; Bhattacharjee, A.; Pang, B.; Nguyen, T.; Williams, J.~J.; Kumar, H.; Popowski, S.; Beltzer, M.; Karr, C.~J.; et~al. 2023.
\newblock A text messaging intervention to support the mental health of young adults: User engagement and feedback from a field trial of an intervention prototype.
\newblock \emph{Internet Interventions}, 100667.

\bibitem[{Kumar et~al.(2022)Kumar, Musabirov, Shi, Lauzon, Choy, Gross, Kulzhabayeva, and Williams}]{kumar2022exploring}
Kumar, H.; Musabirov, I.; Shi, J.; Lauzon, A.; Choy, K.~K.; Gross, O.; Kulzhabayeva, D.; and Williams, J.~J. 2022.
\newblock Exploring the design of prompts for applying gpt-3 based chatbots: A mental wellbeing case study on mechanical turk.
\newblock \emph{arXiv preprint arXiv:2209.11344}.

\bibitem[{Kumar et~al.(2023{\natexlab{a}})Kumar, Wang, Shi, Musabirov, Farb, and Williams}]{kumar2023exploring}
Kumar, H.; Wang, Y.; Shi, J.; Musabirov, I.; Farb, N.~A.; and Williams, J.~J. 2023{\natexlab{a}}.
\newblock Exploring the Use of Large Language Models for Improving the Awareness of Mindfulness.
\newblock In \emph{Extended Abstracts of the 2023 CHI Conference on Human Factors in Computing Systems}, 1--7.

\bibitem[{Kumar et~al.(2023{\natexlab{b}})Kumar, Yu, Chung, Shi, and Williams}]{kumar2023chatbot}
Kumar, H.; Yu, K.; Chung, A.; Shi, J.; and Williams, J.~J. 2023{\natexlab{b}}.
\newblock Exploring The Potential of Chatbots to Provide Mental Well-Being Support for Computer Science Students.
\newblock In \emph{Proceedings of the 54th ACM Technical Symposium on Computer Science Education V. 2}, SIGCSE 2023, 1339. New York, NY, USA: Association for Computing Machinery.
\newblock ISBN 9781450394338.

\bibitem[{Li et~al.(2010)Li, Chu, Langford, and Schapire}]{li2010contextual}
Li, L.; Chu, W.; Langford, J.; and Schapire, R.~E. 2010.
\newblock A contextual-bandit approach to personalized news article recommendation.
\newblock In \emph{Proceedings of the 19th international conference on World wide web}, 661--670.

\bibitem[{Li et~al.(2021)Li, Nogas, Song, Kumar, Durand, Rafferty, Deliu, Villar, and Williams}]{postdiff}
Li, T.; Nogas, J.; Song, H.; Kumar, H.; Durand, A.; Rafferty, A.; Deliu, N.; Villar, S.~S.; and Williams, J.~J. 2021.
\newblock Algorithms for Adaptive Experiments that Trade-off Statistical Analysis with Reward: Combining Uniform Random Assignment and Reward Maximization.

\bibitem[{Meyerhoff et~al.(2022{\natexlab{a}})Meyerhoff, Kornfield, Mohr, and Reddy}]{meyerhoff2022meeting}
Meyerhoff, J.; Kornfield, R.; Mohr, D.~C.; and Reddy, M. 2022{\natexlab{a}}.
\newblock Meeting Young Adults' Social Support Needs across the Health Behavior Change Journey: Implications for Digital Mental Health Tools.
\newblock \emph{Proceedings of the ACM on Human-Computer Interaction}, 6(CSCW2): 1--33.

\bibitem[{Meyerhoff et~al.(2022{\natexlab{b}})Meyerhoff, Nguyen, Karr, Reddy, Williams, Bhattacharjee, Mohr, and Kornfield}]{meyerhoff2022system}
Meyerhoff, J.; Nguyen, T.; Karr, C.~J.; Reddy, M.; Williams, J.~J.; Bhattacharjee, A.; Mohr, D.~C.; and Kornfield, R. 2022{\natexlab{b}}.
\newblock System design of a text messaging program to support the mental health needs of non-treatment seeking young adults.
\newblock \emph{Procedia Computer Science}, 206: 68--80.

\bibitem[{Rafferty et~al.(2019)Rafferty, Ying, Williams et~al.}]{rafferty2019statistical}
Rafferty, A.; Ying, H.; Williams, J.; et~al. 2019.
\newblock Statistical consequences of using multi-armed bandits to conduct adaptive educational experiments.
\newblock \emph{Journal of Educational Data Mining}, 11(1): 47--79.

\bibitem[{Rainie and Zickuhr(2015)}]{rainie_chapter_2015}
Rainie, L.; and Zickuhr, K. 2015.
\newblock Chapter 1: {Always} on {Connectivity}.

\bibitem[{Reza et~al.(2021)Reza, Kim, Bhattacharjee, Rafferty, and Williams}]{reza2021mooclet}
Reza, M.; Kim, J.; Bhattacharjee, A.; Rafferty, A.~N.; and Williams, J.~J. 2021.
\newblock The MOOClet Framework: Unifying Experimentation, Dynamic Improvement, and Personalization in Online Courses.
\newblock In \emph{Proceedings of the Eighth ACM Conference on Learning@ Scale}, 15--26.

\bibitem[{Russo(2020)}]{russo_2020}
Russo, D. 2020.
\newblock Simple bayesian algorithms for best-arm identification.
\newblock \emph{Operations Research}, 68(6): 1625–1647.

\bibitem[{Torous et~al.(2020)Torous, Myrick, Rauseo-Ricupero, Firth et~al.}]{torous2020digital}
Torous, J.; Myrick, K.~J.; Rauseo-Ricupero, N.; Firth, J.; et~al. 2020.
\newblock Digital mental health and COVID-19: using technology today to accelerate the curve on access and quality tomorrow.
\newblock \emph{JMIR mental health}, 7(3): e18848.

\bibitem[{Wang et~al.(2022)Wang, Ma, Zhang*, Liu, and Ma}]{wang2022survey}
Wang, Y.; Ma, W.; Zhang*, M.; Liu, Y.; and Ma, S. 2022.
\newblock A survey on the fairness of recommender systems.
\newblock \emph{ACM Journal of the ACM (JACM)}.

\bibitem[{Williams et~al.(2021)Williams, Nogas, Deliu, Shaikh, Villar, Durand, and Rafferty}]{Williams2021-ll}
Williams, J.~J.; Nogas, J.; Deliu, N.; Shaikh, H.; Villar, S.~S.; Durand, A.; and Rafferty, A. 2021.
\newblock Challenges in statistical analysis of data collected by a bandit algorithm: An empirical exploration in applications to adaptively randomized experiments.

\bibitem[{Wykes, Lipshitz, and Schueller(2019)}]{wykes2019towards}
Wykes, T.; Lipshitz, J.; and Schueller, S.~M. 2019.
\newblock Towards the design of ethical standards related to digital mental health and all its applications.
\newblock \emph{Current Treatment Options in Psychiatry}, 6(3): 232--242.

\bibitem[{Xiang et~al.(2022)Xiang, West, Wang, Cui, and Huang}]{xiang2022multi}
Xiang, D.; West, R.; Wang, J.; Cui, X.; and Huang, J. 2022.
\newblock Multi Armed Bandit vs. A/B Tests in E-commerce-Confidence Interval and Hypothesis Test Power Perspectives.
\newblock In \emph{Proceedings of the 28th ACM SIGKDD Conference on Knowledge Discovery and Data Mining}, 4204--4214.

\bibitem[{Yao et~al.(2021)Yao, Brunskill, Pan, Murphy, and Doshi-Velez}]{yao2021power}
Yao, J.; Brunskill, E.; Pan, W.; Murphy, S.; and Doshi-Velez, F. 2021.
\newblock Power Constrained Bandits.
\newblock In \emph{Machine Learning for Healthcare Conference}, 209--259. PMLR.

\bibitem[{Zhan et~al.(2021)Zhan, Hadad, Hirshberg, and Athey}]{zhan2021off}
Zhan, R.; Hadad, V.; Hirshberg, D.~A.; and Athey, S. 2021.
\newblock Off-policy evaluation via adaptive weighting with data from contextual bandits.
\newblock In \emph{Proceedings of the 27th ACM SIGKDD Conference on Knowledge Discovery \& Data Mining}, 2125--2135.

\end{thebibliography}

\appendix

\newpage

\section{System Architecture}
The architecture of the DMH service consists of two systems - the Dialogue System and the ML Personalization System. This was developed by drawing insights from open-source adaptive experiment infrastructures \cite{reza2021mooclet}. The Dialogue System stores all messaging protocols and content. Researchers can use it to schedule message sequences for participants and send them via the Short Message Service (SMS). To account for privacy and data security concerns \cite{wykes2019towards}, all Personally Identifiable Information (PII) of users were anonymized and appropriate Data Use Agreements (DUA) were set up between the collaborating organizations. The Dialogue System and the Personalization System talk to each other through End-to-End API calls for getting arm assignments, updating context values, and updating rewards. The Personalization system is designed to personalize message content, as well as to adapt on average if there is no effect of contextual variables, to users with different policy strategies. For every observation, the system chooses either Uniform Random or contextual MAB assignment (with equal probability) to sample arms (e.g. provide rationale for introducing a psychological strategy or not), to compare both policies side by side. A separate database is designed for the Personalization System for collecting contextual values and rewards without any private information (phone numbers, emails, etc.) from users. In addition, the Personaliztion System updates parameters of contextual MAB algorithms every 5 minutes during the study. The application was deployed through Mental Health America (MHA), which is identified as a mental health advocacy organization. Upon sign-up, users were informed that this is not a crisis service and were given contact information for appropriate resources for 24/7 human staff care support. An automated risk management software based on a set of keyword rules was used. Users who discussed topics such as suicide were immediately sent a text message providing the appropriate resources to access support.

\begin{figure}
    \includegraphics[width=0.5\textwidth]{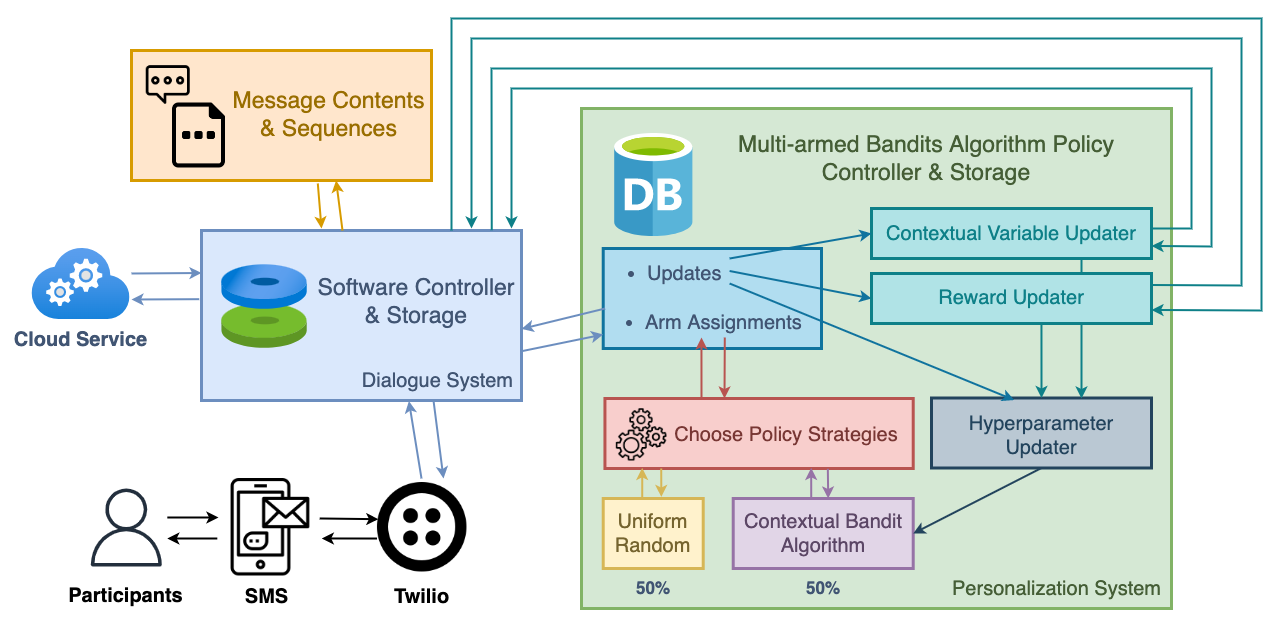}
    \caption{Overall design of the software system. The system comprises of two main components - Dialogue System (to deliver text-messages) and Personalization System.}
    \label{fig:system_stage}
\end{figure}
\label{system*}


\section{Roles and User Interaction flows}

\begin{figure*}
    \includegraphics[width=\textwidth]{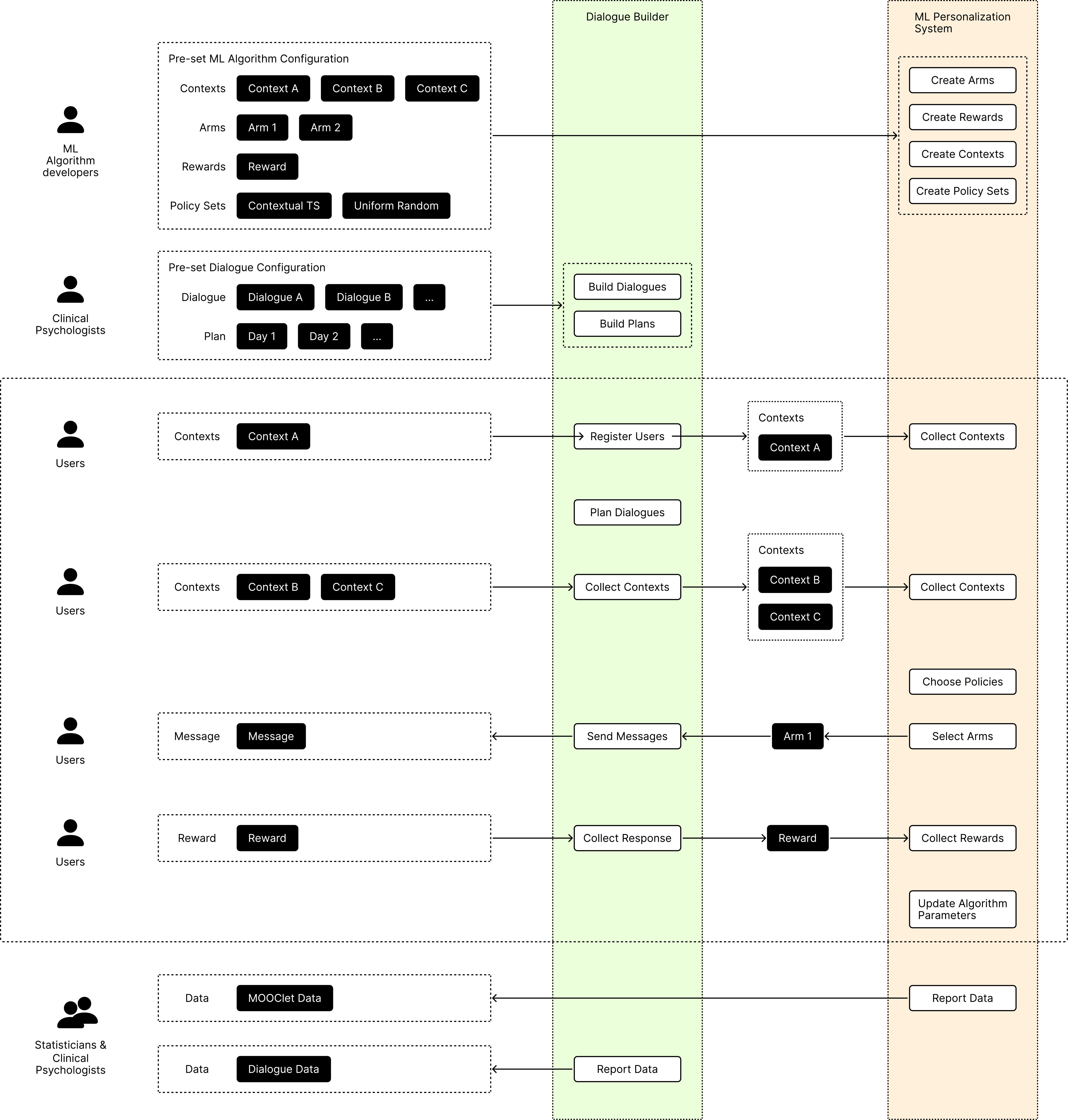}
    \caption{Workflow diagram illustrating the collaboration between ML Algorithm developers, Clinical Psychologists, Users, and Statisticians in creating and refining a personalized dialogue system.}
    \label{fig:user_stage}
\end{figure*}

In Figure \ref{fig:user_stage}, we can see the role of different actors involved in the development, implementation and deployment of the overall system:
\begin{enumerate}
    \item \textbf{ML Algorithm developers}: They are involved in pre-setting A/B Testing policy configuration. Their tasks include defining different contexts such as Context A, Context B, and Context C. They also determine various arms (Arm 1, Arm 2, etc.) and rewards with scales for the system. In addition, they set up policy sets, including contextual TS and Uniform Random.
    \item \textbf{Clinical Psychologists}: They are responsible for pre-setting dialogue configurations. This involves building different dialogues, such as Dialogue A, Dialogue B, etc. They also develop plans for scheduling Dialogues which comprise various days like Day 1, Day 2, etc.
    \item \textbf{Users}: Users have multiple tasks within the system. Initially, they are registered and assigned to certain contexts, like Context A. They are also involved in planning dialogues. As they interact with the system, they provide different contexts (Context B, Context C, etc.), send messages, and receive rewards. The system collects their responses after receiving the messages.
    \item \textbf{Statisticians \& Clinical Psychologists}: Their main role is to analyze the data generated from the system. They look into data from the Personalization System and Dialogue Builder to gain insights and report on the findings.
\end{enumerate}
The entire workflow showcases an interaction between machine learning algorithms, user input, clinical interventions, and data analysis to create a personalized dialogue system.

\section{Sample Messaging Schedule for the Users}
\begin{figure*}
    \includegraphics[width=\textwidth]{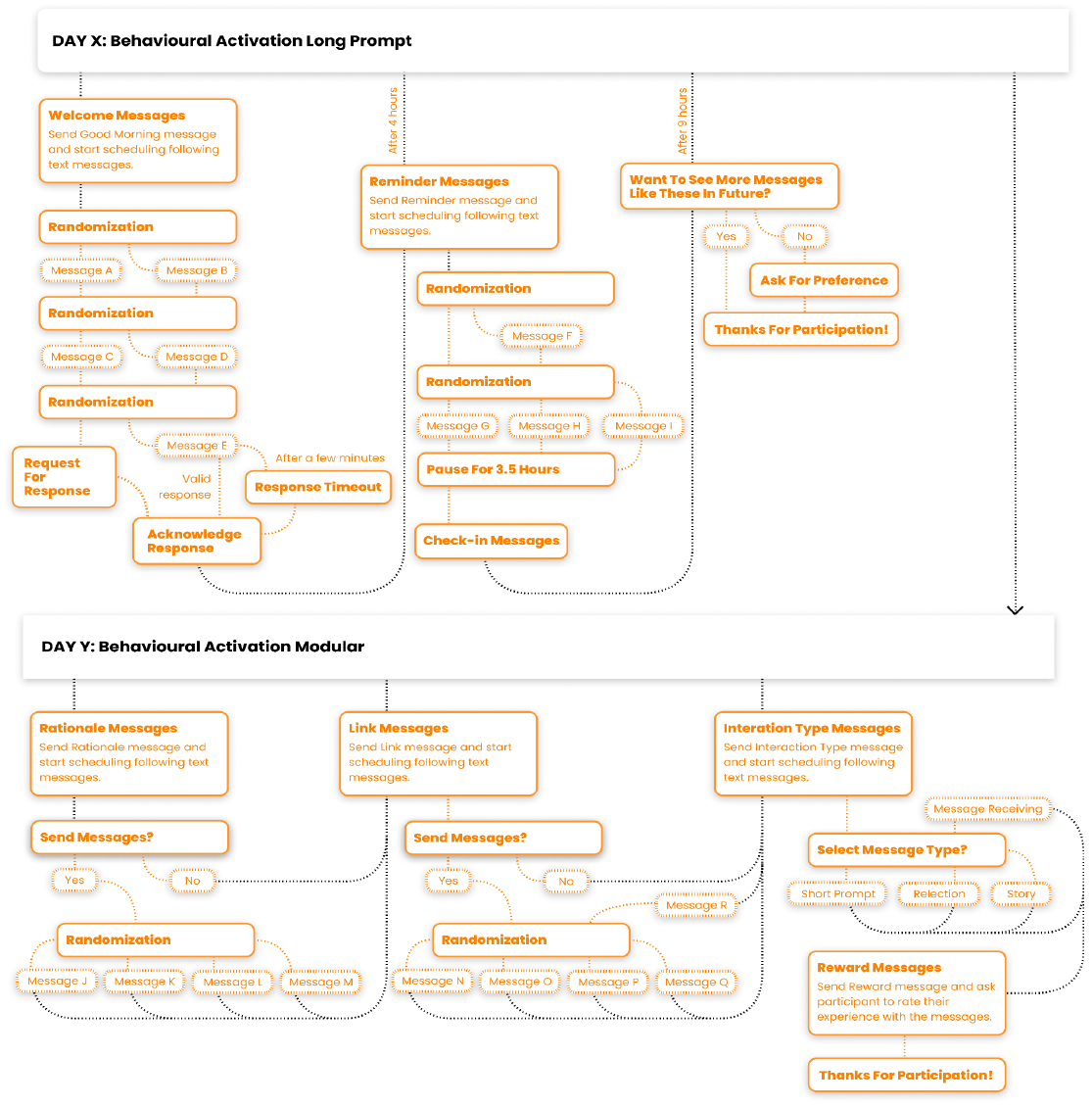}
    \caption{Example sequence of messages received by the user at different points of the study. The diagram showcases various possible randomization points for personalization to make the messages more engaging for the users.}
    \label{fig:protocol}
\end{figure*}

Figure \ref{fig:protocol} shows an example sequence of messages that the user might receive during the course of the intervention. While the focus of this paper was to illustrate the applicability of our system for "Modular" interactions, the study involves longer sequences of messages with various points for personalizing the messages.

\newpage

\newpage

\end{document}